\def\year{2021}\relax
\documentclass[letterpaper]{article} 
\usepackage{aaai21}  
\usepackage{times}  
\usepackage{helvet} 
\usepackage{courier}  
\usepackage[hyphens]{url}  
\usepackage{graphicx} 
\usepackage{natbib}  
\usepackage{caption} 
\title{Clinical Temporal Relation Extraction with Probabilistic Soft Logic Regularization and Global Inference}

\author{
    Yichao Zhou\textsuperscript{\rm 1}, 
    Yu Yan\textsuperscript{\rm 2}, 
    Rujun Han\textsuperscript{\rm 3}, 
    J. Harry Caufield\textsuperscript{\rm 2}, \\
    Kai-Wei Chang\textsuperscript{\rm 1}, 
    Yizhou Sun\textsuperscript{\rm 1}, 
    Peipei Ping\textsuperscript{\rm 2}, 
    and Wei Wang\textsuperscript{\rm 1} \\
}

\affiliations {
    \textsuperscript{\rm 1} Department of Computer Science, University of California, Los Angeles. \\
    \textsuperscript{\rm 2} Departments of Physiology, Medicine and Bioinformatics, University of California, Los Angeles.\\
    \textsuperscript{\rm 3} Department of Computer Science, University of Southern California, Los Angeles. \\
    \{yz, kwchang, yzsun, weiwang\}@cs.ucla.edu, rujunhan@usc.edu, \{yuyan, jcaufield, pping\}@mednet.ucla.edu
}

\usepackage{xspace}
\usepackage{xcolor}
\usepackage{enumitem}
\usepackage{bm}
\usepackage{graphicx}
\usepackage{subcaption}
\usepackage{amsmath}
\usepackage{url}
\usepackage{array,multirow}
\usepackage[utf8]{inputenc}
\usepackage[english]{babel}

\usepackage{amsthm}
\theoremstyle{plain}
\newtheorem{mydef}{Definition}[]
\theoremstyle{remark}
\newtheorem{myex}{Example}[]

\usepackage{amssymb}
\usepackage{adjustbox}
\usepackage[ruled,vlined]{algorithm2e} 

\newcommand{\modelname}{\texttt{CTRL-PG}\xspace}

\newcommand{\modelfullname}{\underline{C}linical \underline{T}emporal \underline{R}e\underline{L}ation Exaction with \underline{P}robabilistic Soft Logic Regularization and \underline{G}lobal Inference\xspace}
\DeclareMathOperator*{\argmax}{argmax}

\usepackage{array,booktabs,ragged2e}
\newcolumntype{R}[1]{>{\RaggedLeft\arraybackslash}p{#1}}

\begin{document}

\maketitle

\begin{abstract}
There has been a steady need in the medical community to precisely extract the temporal relations between clinical events. In particular, temporal information can facilitate a variety of downstream applications such as case report retrieval and medical question answering. However, existing methods either require expensive feature engineering or are incapable of modeling the global relational dependencies among the events. In this paper, we propose \modelfullname (\modelname), a novel method to tackle the problem at the document level. Extensive experiments on two benchmark datasets, I2B2-2012 and TB-Dense, demonstrate that \modelname significantly outperforms baseline methods for temporal relation extraction.   
\end{abstract}
\section{Introduction}

Clinical case reports (CCRs) are written descriptions of the unique aspects of a particular clinical case~\citep{Caban-Martinez2012, caufield2018reference}. 
They are intended to serve as educational aids to science and medicine, as they play an essential role in sharing clinical experiences about atypical disease phenotypes and new therapies~\citep{caufield2018reference}. 
There is a perennial need to automatically and precisely curate the clinical case reports into structured knowledge, i.e. extract important clinical named entities and relationships from the narratives~\cite{aronson2010overview,savova2010mayo,soysal2018clamp,caufield2019comprehensive,alfattni2020extraction}.
This would greatly enable both doctors and patients to retrieve related case reports for reference and provide a certain degree of technical support for resolving public health crises like the recent COVID-19 pandemic. 
Clinical reports describe chronicle events, elucidating a chain of clinical observations and reasoning~\cite{sun2013evaluating,chen2016orderrex}. 
Extracting temporal relations between clinical events is essential for the case report retrieval over the patient chronologies. Besides, medical question answering systems require the precise ordering of clinical events in a time series within each document. 

In this paper, we tackle the temporal relation extraction problem in clinical case reports. 
\begin{figure}[t]
    \centering
    \includegraphics[width=\linewidth]{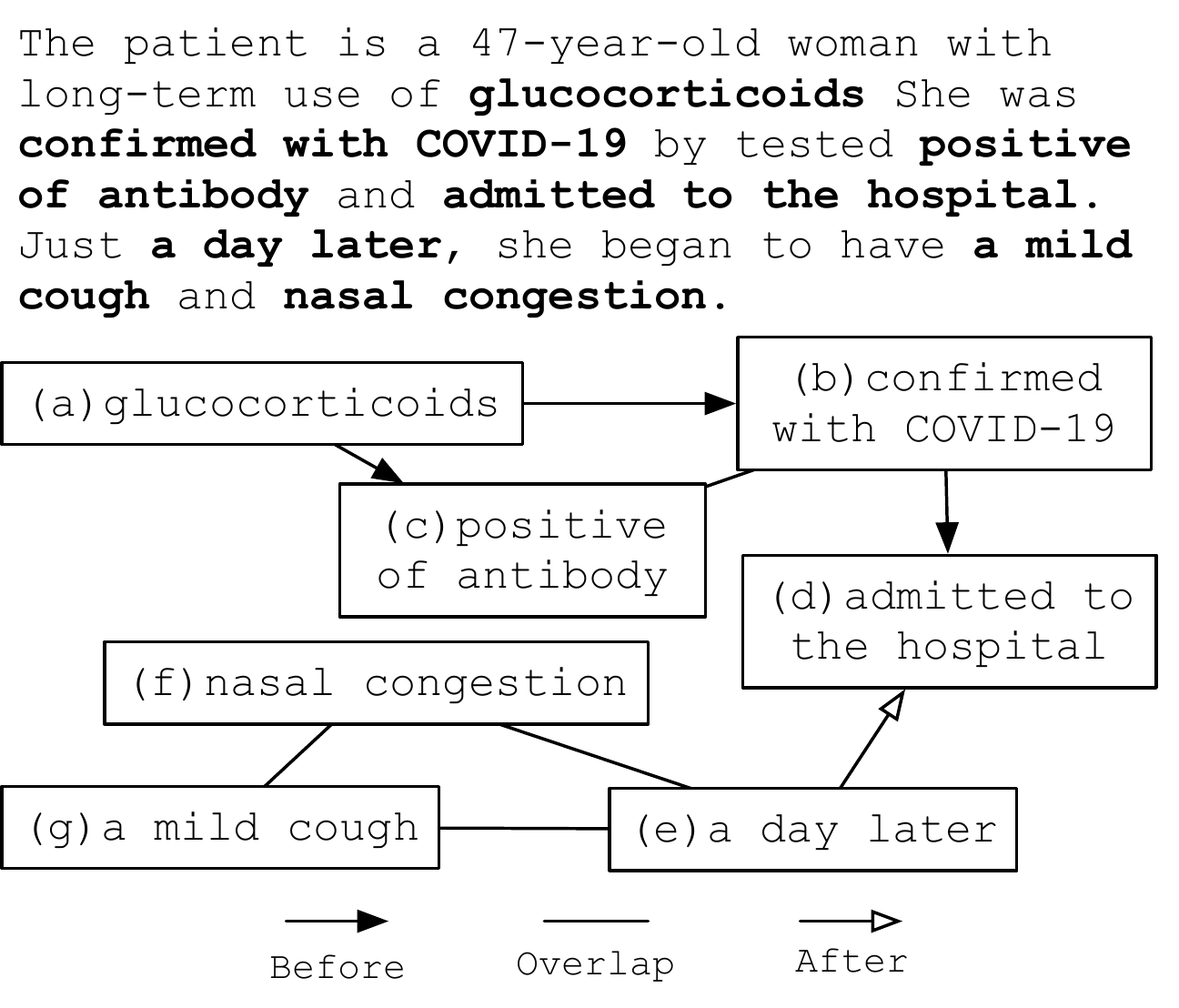}
    \caption{An illustration of a clinical case report with its partial temporal graph where transitivity dependencies exist. 
    }
    \label{fig:case_report.pdf}
\end{figure}
Figure~\ref{fig:case_report.pdf} illustrates a paragraph from a typical CCR document with three common types of temporal relations, ``Before'', ``After'', and ``Overlap''. \textit{Glucocortocoids} was described as the medicine history of this patient, which happened before \textit{confirmed with COVID-19} and \textit{positive of antibody}.  
An ``Overlap'' temporal relation exists between \textit{nasal congestion} and \textit{a mild cough}. 
We consider the aforementioned clinical concepts as events, while regarding \textit{a day later} as a time expression. A temporal relation may exist between event and event ($\texttt{E-E}$), event and time expression ($\texttt{E-T}$) or time expression and time expression ($\texttt{T-T}$). 

There is a consensus within the clinical community regarding the difficulty of temporal information extraction, due to the high demand for domain knowledge and high complexity of clinical language representations
~\cite{galvan2018investigating}. 
\citet{meng-rumshisky-2018-context,lee-etal-2016-uthealth} apply machine learning models with lexical, syntactic features, or pre-trained word representations to tackle the problem but neglect the strong dependencies between narrative containment and temporal order, thus predicting inconsistent output labels and garbled time-lines~\cite{leeuwenberg-moens-2017-structured}. 

The dependency is the key enabler of classifying the temporal relations. 
For instance in Figure~\ref{fig:case_report.pdf}, given that \texttt{b} happened before \texttt{d}, \texttt{e} happened after \texttt{d} and \texttt{e} happened simultaneously with \texttt{f} , we can infer according to the temporal transitivity rule that \texttt{b} was before \texttt{f}. 
Some recent studies~\cite{leeuwenberg-moens-2017-structured,ning-etal-2017-structured,han2020knowledge} 
convert the task to a structured prediction problem 
and solve it with Maximum a posteriori Inference. 
Integer Linear Programming (ILP) with hard constraints is deployed for optimization, which however needs an off-the-shelf solver to tackle the NP-hard optimization problem and can only approximate the optimum via relaxation. Besides, globally inferring the relations at the document level would also be intractable for them due to the high complexity and low scalability~\cite{bach2017hinge}. 

Recently, some researchers~\cite{deng-wiebe-2015-joint,chen2019embedding,hu-etal-2016-harnessing} have explored 
Probabilistic Soft Logic (PSL)~\cite{bach2017hinge} to tackle the structured prediction problem. 
Inspired by them, we propose to leverage the PSL rules to model 
relation extraction more flexibly and efficiently.
In specific, we summarize common transitivity and symmetry patterns of temporal relations as PSL rules and penalize the training instances that violate any of those rules. Different from ILP solutions, no off-the-shelf solver is required and the algorithm conducts the training process with linear time complexity. Besides, logical propositions in PSL can be interpreted not just as $true$ or $false$, but as continuously valued in the $[0, 1]$ interval. We also propose a simple but effective time-anchored global temporal inference algorithm to classify the relations at the document level. With such a mechanism, we can easily verify some relations, such as the relation between \texttt{b} and \texttt{f}, with long-term dependencies which are intractable with existing approaches.     
As a summary, our main contributions are list as follows:
\begin{itemize}
    \item To the best of our knowledge, this is the first work to formulate the probabilistic soft logic rules of temporal dependencies as a regularization term to jointly learn a relation classification model,
    \item We show the efficacy of globally inferring the temporal relations with the time graphs,
    \item We release the codes\footnote{\url{https://github.com/yuyanislearning/CTRL-PG}.} to facilitate further developments by the research community.
\end{itemize}

Next, we give the problem definition and explain how we leverage PSL rules to model the temporal dependencies. We then describe the overall architecture of our clinical temporal relation extraction model, \modelname and show the extensive experimental results in the following sections.  

\section{Preliminaries}\label{sec:psl}
\subsection{Problem Statement}
Document $D$ contains sequences $[s_1,s_2, ..., s_M]$ and named entities $x_i \in \mathcal{E} \bigcup \mathcal{T}, 1 \leq i \leq N$, where $M,N$ are the total number of sequences and entities in $D$. $\mathcal{E}$ and $\mathcal{T}$ represent the set of events and time expressions, respectively. There is a potential temporal relation between any pair of annotated named entities $(x_j, x_k)$, where $1\leq j,k \leq N$. Formally, the task is modeled as a classification problem with a set of temporal relation types $\mathcal{Y}$. 
Given a sequence $s_i$ together with two named entities $x_{i,1},x_{i,2}$ included, we predict the temporal relation $y_i\in \mathcal{Y}$ from $x_{i,1}$ to $x_{i,2}$. 
In practice, we create a triplet with three pairs of entities to be one training instance $\mathcal{I}$, to enable the PSL rule grounding, as explained in the following section.

\subsection{Probabilistic Soft Logic and Temporal Dependencies in Clinical Narratives}\label{sec:psld}
Here, we introduce some concepts and notations for the language PSL and illustrate how PSL is applicable to define templates for temporal dependencies and to help jointly learn a relation classifier. 

\begin{mydef}
A \textbf{predicate} $\tilde{p}$ is a relation defined by a unique identifier and an \textbf{atom} $\tilde{l}$ is a predicate combined with a sequence of terms of length equal to the predicate’s argument number. Atoms in PSL take on continuous values in the unit interval $[0, 1]$. 
\end{mydef}
\begin{myex}
$\underline{\text{Before}}/2$ indicates a predicate taking two arguments, and the atom $\underline{\text{Before}}(A,B)$ represents whether $A$ happens before $B$.
\end{myex}
\begin{mydef}\label{def:rule}
A \textbf{PSL rule} $\tilde{r}$ is a disjunctive clause of atoms or negative atoms:
\begin{equation}
    \eta_r: T_1 \land T_2 \land ... \land T_{m} \rightarrow H_1 \lor H_2 \lor ... \lor H_{n},
\end{equation}
where $T_1,T_2,...,T_m,H_1,H_2,...,H_n$ are atoms or negative atoms. 
\end{mydef}
We name  $T_1,T_2,...,T_m$ as $r_{body}$ and $H_1,H_2,...,H_n$ as $r_{head}$.  $\eta_r\in [0,1]$ is the weight of the rule $r$, denoting the prior confidence of this rule. To the opposite, an unweighted PSL rule is to describe a constraint that is always true. The unweighted logical clauses in Table~\ref{tab:psl} describe the common temporal transitivity and symmetry dependencies we summarize from the clinical narratives.
\begin{table}[t]
    \centering
    \resizebox{\linewidth}{!}{
    \begin{tabular}{|c|c|}
    \hline
        Abbrev. & PSL rules  \\
        \hline
        \hline
        \multicolumn{2}{|l|}{Transitivity Dependencies} \\
        \hline
        BBB & Before$(A,B)$ $\land$ Before$(B,C)$ $\rightarrow$ Before$(A,C)$ \\
        BOB & Before$(A,B)$ $\land$ Overlap$(B,C)$ $\rightarrow$ Before$(A,C)$ \\
        OBB & Overlap$(A,B)$ $\land$ Before$(B,C)$ $\rightarrow$ Before$(A,C)$ \\
        OOO & Overlap$(A,B)$ $\land$ Overlap$(B,C)$ $\rightarrow$ Overlap$(A,C)$ \\
        AAA & After$(A,B)$ $\land$ After$(B,C)$ $\rightarrow$ After$(A,C)$ \\
        AOA & After$(A,B)$ $\land$ Overlap$(B,C)$ $\rightarrow$ After$(A,C)$ \\
        OAA & Overlap$(A,B)$ $\land$ After$(B,C)$ $\rightarrow$ After$(A,C)$ \\
        \hline
        \hline
        \multicolumn{2}{|l|}{Symmetry Dependencies} \\
        \hline
        BA & Before$(A,B)$ $\rightarrow$ After$(B,A)$ \\
        AB & After$(A,B)$ $\rightarrow$ Before$(B,A)$\\
        OO & Overlap$(A,B)$ $\rightarrow$ Overlap$(B,A)$ \\
    \hline
    \end{tabular}}
    \caption{Temporal transitivity and symmetry PSL rules $\mathcal{R}$. $A,B,C$ are three terms representing either events or time expressions.}
    \label{tab:psl}
\end{table}
\begin{mydef}
The \textbf{ground atom} $l$ and \textbf{ground rule} $r$ are particular variable instantiation of some atom $\tilde{l}$ and rule $\tilde{r}$, respectively. 
\end{mydef}
\begin{myex}\label{ex:rule}
That \underline{Overlap} (\texttt{e}, \texttt{f}) $\land$ \underline{Overlap} (\texttt{f}, \texttt{g}) $\rightarrow$ \underline{Overlap} (\texttt{e}, \texttt{g}) from Figure~\ref{fig:case_report.pdf} is a ground rule composed of three ground atoms, denoted as $l_1,l_2$, and $l_3$, respectively. It is grounded from the OOO rule, as shown in Table~\ref{tab:psl}. 
\end{myex}
\begin{mydef}
The interpretation $I(l)$ denotes the soft truth value of an atom $l$. 
\end{mydef}
\begin{mydef}
Łukasiewicz t-norm~\cite{klir1995fuzzy} is used to define the basic
logical operations in PSL, including logical conjunction ($\land$), disjunction ($\lor$), and negation ($\neg$):
\begin{align}
&    I(l_1 \land l_2) = \max\{I(l_1) + I(l_2) - 1, 0 \}\label{eq:luk} \\
&    I(l_1 \lor l_2) = \min\{I(l_1) + I(l_2), 1 \} \\
&    I(\neg l_1) = 1 - I(l_1)
\end{align}
\end{mydef}
The PSL rule in Definition~\ref{def:rule} can also be represented as: 
\begin{align*}
    I(r_{body}\rightarrow r_{head}) = I(\neg r_{body} \lor r_{head}),  
\end{align*}
so we can induce the distance to satisfaction for rule $r$.
\begin{mydef}
The \textbf{distance to satisfaction} $d_r(I)$ of rule $r$ under an interpretation $I$ is defined as:
\begin{align}\label{eq:dis}
    d_r(I) = \max\{0, I(r_{body})-I(r_{head})\}
\end{align}
\end{mydef}
PSL program determines a rule $r$ as satisfied when the truth value of  $I(r_{head})-I(r_{body})\geq 0$.
\begin{myex}
Given that $I(l_1) = 0.7, I(l_2) = 0.8, $ and $I(l_3) = 0.3$, we can compute the distance according to Equation~\eqref{eq:luk}-\eqref{eq:dis}:
\begin{align*}
 &   d_r = \max\{0, I(l_1 \land l_2) - I(l_3)\} \\
 &       = \max\{0, 0.7 + 0.8 -1 - I(l_3)\} \\
 &       = \max\{0, 0.5 -0.3\} \\
 &       = 0.2
\end{align*}
\end{myex}
This equation indicates that the ground rule in Example~\ref{ex:rule}  is completely satisfied when $I(l_3)$ is above $0.5$. Otherwise, a penalty factor will be raised ($0.2$ in this case). When $I(l_3)$ is under $0.5$, the smaller $I(l_3)$ is, the larger penalty we have. 
In short, we compute the distance to satisfaction for each ground rule as a loss regularization term to jointly learn a relation classification model. 
We finally use the smallest one as the penalty because we only need one of the rules to be satisfied.

\begin{figure*}[!t]
    \centering
    \includegraphics[width=\linewidth]{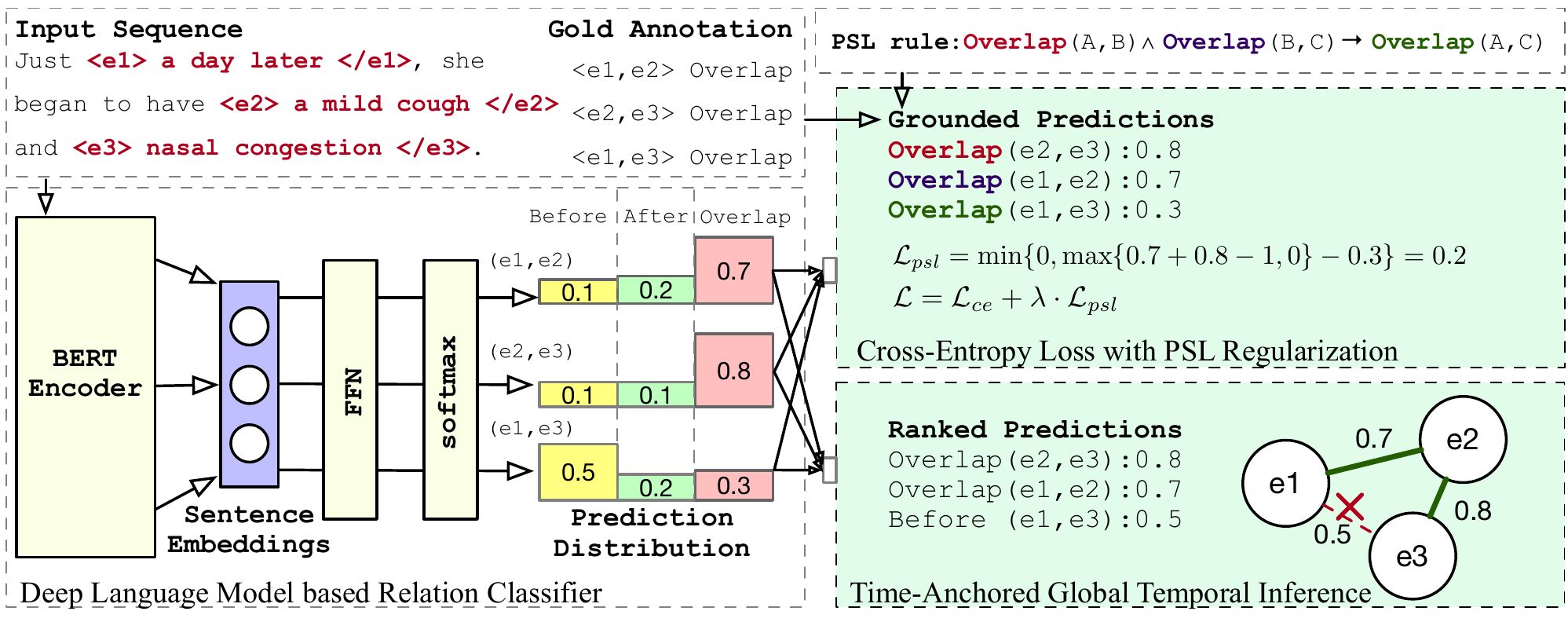}
    \caption{The overall architecture of \modelname. We leverage the cross-entropy function and PSL regularization to jointly train a relation classifier with the sentence embeddings from a deep language model and perform global temporal inference with a time graph. 
    We acquire the prediction probability distributions from a deep language model based relation classifier to compute the distance loss $\mathcal{L}_{psl}$ in the PSL module and time graph construction in the global inference module. }
    \label{fig:framework}
\end{figure*}

\section{Clinical Temporal Relation Extraction}\label{sec:model}
Figure~\ref{fig:framework} shows the overall framework of the proposed \modelname model. The framework consists of three components, (i) a temporal relation classifier composed of a deep language encoder and a Feed-Forward Network (FFN), (ii) a Cross-Entropy loss function with PSL regularization, and (iii) a time-anchored global temporal inference module. We will introduce the details of the three modules in the following subsections.

\subsection{Temporal Relation Classifier}
The context is essential for capturing the syntactic and semantic features of each word in a sequence. Hence, we propose to apply the contextualized language model, BERT~\cite{devlin2018bert}, to derive the sentence representation $v_i$ of $d_s$-dimension to encode the input sequence $s_i$ including two marked named entities $x_{i,1}, x_{i,2}$ from the instance $\mathcal{I}$, where $i\in\{1,2,3\}$. We group three sequences together to facilitate the computation of regularization term introduced in the next subsection.

By feeding the sentence embedding $v_i$ to a layer of FFN, we can predict the relation type $\hat{y_i}$ with the softmax function:
\begin{align}
&    \hat{y}_i  = \argmax_{y \in \mathcal{Y}} \; \mathbb{P} (y|s_i)\\
&    \mathbb{P}(y|s_i) = \text{softmax}( W_f \cdot v_i + b_f), 
\end{align}
where $W_f$ and $b_f$ are the weights and bias in the FFN layer. 

To learn the relation classification model, we first compute a loss with the Cross-Entropy objective for each instance $\mathcal{I}$:
\begin{align}\label{eq:ce}
& \mathcal{L}_{ce} = -\sum_{i\in\{1,2,3\}}\sum_{y\in\mathcal{Y}} y\log\mathbb{P}(y|s_i)
\end{align} 

\subsection{Learning with Probabilistic Soft Logic Regularization}\label{sec:pslr}
We also aim to minimize the distance to rule satisfaction for each instance. We compute the distance with function $\mathcal{F}(\cdot,\cdot)$, as described in Algorithm~\ref{algo:PSL},
by finding the minimum of all possible PSL rule grounding results, i.e., when one PSL rule is satisfied, $\mathcal{F(\cdot,\cdot)}$ should return $0$. 
In specific, we first ground the three relation predictions $\hat{y}_{i}$ with potential PSL rules.
We then incorporate Equation~\eqref{eq:luk}-\eqref{eq:dis} for distance computation. The prediction probabilities are regarded as the interpretation of the ground atoms $l_i$.  
If none of the rules can be grounded, the distance will be set as 0.    
\begin{algorithm}[t]
\SetAlgoLined
\textbf{Input:} PSL Rules $\mathcal{R}$, Prediction $\hat{y}_i$, and Probability $\mathbb{P}(y|s_i)$, $i=\{1,2,3\}$\;
\textbf{Output:} Distance $d_r$\;
    Set $d_r=1$; $d_t=0$; IsGround $=false$\; 
    \For{each $l_1\land l_2 \to l_3 \in \mathcal{R}$}{
      \uIf{$\hat{y}_{1}$ matches $l_1$ and $\hat{y}_{2}$ matches $l_2$}{
        Determine $\bar{y}_{3}$ with $l_3$\;
        $d_t \leftarrow \max\{\mathbb{P}(y=\hat{y}_{1}|s_1) + \mathbb{P}(y=\hat{y}_{2}|s_2) - 1, 0\}$\;
        $d_t \leftarrow \max\{d_t - \mathbb{P}(y=\bar{y}_{3}|s_3) , 0\}$\;
        $d_r \leftarrow \min\{d_r, d_t\}$\;
        IsGround $\leftarrow true$\;
      }
      
    }
    \uIf{IsGround $==false$}{$d_r \leftarrow 0$\;}
\caption{Function $\mathcal{F}$ for PSL Rule Grounding and Distance Calculation.}
\label{algo:PSL}
\end{algorithm}
Then, we formulate the distance to satisfaction as a regularization term to penalize the predictions that violate any PSL rule:
\begin{align}\label{eq:psl}
& \mathcal{L}_{psl} = \mathcal{F}(\mathcal{R}; \{(\mathbb{P}(y|s_i),\hat{y}_i)\}), i=\{1,2,3\}
\end{align}
and finalize the loss function by summing up \eqref{eq:ce} and \eqref{eq:psl}:
\begin{align}\label{eq:loss}
& \mathcal{L} = \mathcal{L}_{ce} + \lambda\cdot \mathcal{L}_{psl},
\end{align}
where $\lambda$ is a hyperparameter as the weight for PSL regularization term.
We apply gradient descent to minimize the loss function~\eqref{eq:loss} and to update the parameters of our model. 

\subsection{Global Temporal Inference}
In the inference stage, we leverage the Timegraph algorithm~\cite{miller1990time} to resolve the conflicts in the temporal relation predictions $\bm{\hat{y}}$. 
Timegraph is a widely used algorithm of time complexity $\mathcal{O}(v+e)$ for deriving the temporal relation for any two nodes in a connected graph, where $v$ and $e$ denote the numbers of nodes and edges. Nodes and edges represent the named entities and temporal relations, respectively. 
Our goal is to construct a conflict-free time graph $\mathcal{G}$ for each document $D$ through a greedy Check-And-Add process, described as $4$ steps in Algorithm~\ref{algo:gri}. 
Intuitively, we want to rely on some trustworthy edges to resolve the conflicts in the time graph with the transitivity and symmetry dependencies listed in Table~\ref{tab:psl}. As illustrated in Figure~\ref{fig:framework}, the probabilities of predictions \underline{Overlap}(e1, e2) and  \underline{Overlap}(e2, e3) are $0.7$ and $0.8$, which are higher than that of \underline{Overlap}(e1, e3). When we trust the first two predictions, the third prediction could be neglected considering the relation between e1 and e3 can already be inferred with the transitivity dependency. In this way, the predicting mistakes with low confidence scores can be ruled out, leading to better model performance in the closure evaluation.

We believe that the relations between time expressions are the easiest ones to predict. For example, the ground atom \underline{Before} (\textit{06-15-91}, \textit{July 1st 1991}) is obviously $true$. Therefore, we try to build up a base time graph on top of the relations of type \texttt{T-T}. Next, we rank the rest of the predictions according to their probabilities in decreasing order and then check whether each of the predictions is inconsistent with the current time graph iteratively. The relation will be dropped if it raises a conflict, otherwise added to the graph as a new edge.   

\begin{algorithm}[t]
\SetAlgoLined
Step 1: Predict temporal relations $P_1$ on pairs of the time expressions \texttt{T-T}\;
Step 2: Construct a time graph $\mathcal{G}$ with $P_1$\;
Step 3: Rank all other predictions $P_2$ on the relations of type \texttt{E-E} and \texttt{E-T} according to the predicting probabilities in decreasing order, naming $P_2^{ranked}$\;
Step 4: \\ 
\For{each $p$ in $P_2^{ranked}$}{
    Apply Timegraph algorithm to check the conflict between $p$ and $\mathcal{G}$\;
    \uIf{there exists a conflict}{Drop $p$\;}
    \Else{Add the edge $p$ to $\mathcal{G}$\;}
}
\caption{Check-And-Add Process for Constructing a Conflict-free Time Graph $\mathcal{G}$}
\label{algo:gri}
\end{algorithm}

\section{Experiments}\label{sec:exp}
In this section, we develop experiments on two benchmark datasets to prove the effectiveness of both PSL regularization and global temporal inference. We also discuss the limitation and perform error analyses for \modelname.

\begin{table}[ht!]
    \centering
    \resizebox{.95\linewidth}{!}{
    \begin{tabular}{|c|l|r|r|r|}
    \hline
        \multicolumn{2}{|c|}{Dataset} & Train & Dev & Test \\
        \hline
        \hline
        \multirow{2}{*}{I2B2-2012} & \# doc & 181 & 9 & 120 \\
        \cline{2-5}
        & \# relation & 29,736 & 1,165 & 24,971 \\ 
        \hline
        \multirow{2}{*}{TB-Dense} & \# doc & 22 & 5 & 9 \\
        \cline{2-5}
        & \# relation & 4,032 & 629 & 1,427 \\ 
        \hline
    \end{tabular}}
    \caption{Dataset Statistics.}
    \label{tab:data}
\end{table}

\subsection{Datasets}
Experiments are conducted on I2B2-2012 and TB-Dense datasets and an overview of the data statistics is shown in Table~\ref{tab:data}. The datasets have diverse annotation densities and instance numbers.

\noindent \textbf{I2B2-2012.} The I2B2-2012 challenge corpus~\cite{sun2013evaluating} consists of 310 discharge summaries. Two categories of temporal relations, \texttt{E-T} and \texttt{E-E}, were annotated in each document. 
Three temporal relations\footnote{\citet{sun2013evaluating} merged 7 original temporal relations to 3 to increase Inter-annotator agreement.}, \underline{Before}, \underline{After}, and \underline{Overlap}, were used. I2B2-2012 has a relatively low annotation density\footnote{Annotation density denotes the percentage of annotated pairs of event/time expressions.}, which is $0.21$. 


\noindent \textbf{TB-Dense.} To prove that our PSL regularization is a generic algorithm and can be easily adapted to other domains, we also test it on the TB-dense~\cite{cassidy-etal-2014-annotation} dataset, which is based on TimeBank News Corpus~\cite{pustejovsky2003timebank}. Annotators were required to label all pairs of events/times in a given window to address the sparse annotation issue in the original data. Thus the annotation density is 1. This dataset has six relation types, \underline{Simultaneous}, \underline{Before}, \underline{After}, \underline{Includes}, \underline{Is\_Include}, and \underline{Vague}.

\subsection{Baseline Models}
We employ different baseline models for the two datasets to compare our method with the SOTA models in both clinical and news domains.  
\\

\noindent \textbf{I2B2-2012} (1) Feature-engineering based statistic models from I2B2-2012 challenge, \texttt{MaxEnt-SVM}~\cite{xu2013end} incorporating Maximum Entropy with Support Vector Machine (SVM), \texttt{CRF-SVM}~\cite{tang2013hybrid} using Conditional Random Fields and SVM, \texttt{RULE-SVM}~\cite{nikfarjam2013towards} relying on rule-based algorithms; (2) Neural network based model,  \texttt{RNN-ATT}~\cite{liu-etal-2019-attention}, which applies Recurrent Neural Network plus attention mechanism; (3) Structured Prediction method, \texttt{SP-ILP}~\cite{han-etal-2019-deep,leeuwenberg-moens-2017-structured} leveraging the ILP optimization; (4) Basic version of our model, \texttt{CTRL}, which only fine-tunes a BERT-BASE~\cite{devlin2018bert} language model with one layer of FFN, similar to the implementations in \citet{lin-etal-2019-bert,guan2020robustly}.
\\

\noindent \textbf{TB-Dense.} (1) \texttt{CAEVO}~\cite{chambers2014dense} with a cascade of rule-based classifiers; (2) \texttt{LSTM-DP}~\cite{cheng-miyao-2017-classifying} using LSTM-based network and cross-sentence dependency paths; (3) \texttt{GCL}~\cite{meng-rumshisky-2018-context} incorporating LSTM-based network with discourse-level contexts; (4) \texttt{SP-ILP} and \texttt{CTRL}, same as the baselines for I2B2-2012. Note that the results of \texttt{CAEVO}, \texttt{LSTM-DP}, \texttt{GCL}, and \texttt{SP-ILP} are collected from \citet{han-etal-2019-deep}.  

\subsection{Evaluation Metrics}
To be consistent with previous work for a fair comparison, we adopt two different evaluation metrics. For TB-Dense dataset, we compute the Precision, Recall, and Micro-average F1 scores. Following \cite{han-etal-2019-joint,meng-rumshisky-2018-context}, we only predict the \texttt{E-E} relations and exclude all other relations from evaluation. Note that Micro-averaging in a multi-class setting will lead to the same value for Precision, Recall, and F1.
For I2B2-2012, we leverage the TempEval evaluation metrics used by the official challenge~\cite{sun2013evaluating}, which also calculates the Precision, Recall, and Micro-average F1 scores. 
This evaluation metrics differ from the standard F1 used for TB-Dense in a way that it computes the Precision by verifying each prediction in the closure of the ground truths and computes the Recall by verifying each ground truth in the closure of the predictions. We explore all types of temporal relations in I2B2-2012 dataset.

\subsection{Implementation Details}
In the framework of \modelname, any contextualized word embedding method, such as BERT~\cite{devlin2018bert}, ELMo~\cite{peters2018deep}, and RoBERTa~\cite{liu2019roberta}, can be utilized. We choose BERT~\cite{devlin2018bert} to derive contextualized sentence embeddings without loss of generality. BERT adds a special token \texttt{[CLS]} at the beginning of each tokenized sequence and learns an embedding vector for it. We follow the experimental settings in \cite{devlin2018bert} to use $12$ Transformer layers and attention heads and set the embedding size $d_s$ as 768. 
The \modelname is implemented in PyTorch and we use the fused Adam optimizer~\cite{kingma2014adam} to optimize the parameters. We follow the experimental settings in \cite{devlin2018bert} to set the dropout rate, and batch size as $10^{-1}$ and 8. We perform grid search for the initial learning rate from a range of $\{1\times 10^{-5},2\times 10^{-5},4\times 10^{-5},8\times 10^{-5}\}$ and finally select $2\times 10^{-5}$ for both datasets. 
We train 10 epochs for each experiment on two datasets, which can all be completed within 2 hours on single DGX1 Nvidia GPU. 

\begin{figure}[t]
    \centering
    \includegraphics[width=.90\linewidth]{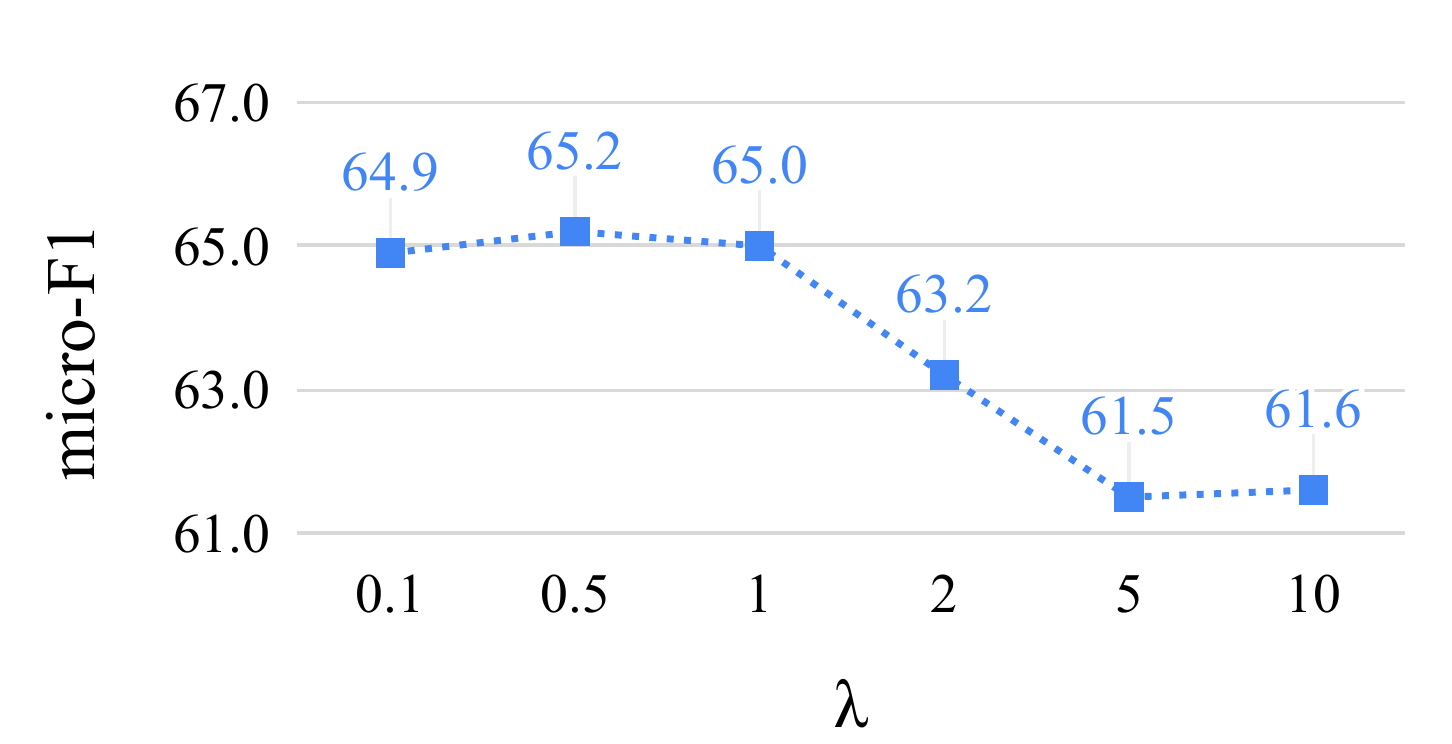}
    \caption{Hyperparameter Search for $\lambda$ on TB-Dense dataset.}
    \label{fig:hyper}
\end{figure} 

To tune the hyperparameters, we search the PSL regularization term $\lambda$ from $\{0.1, 0.5 , 1, 2, 5, 10\}$ as shown in Figure~\ref{fig:hyper}. For I2B2-2012 and TB-Dense datasets, we set $\lambda$ as $5$ and $0.5$, respectively. The hyperparameters are selected by observing the best F1 performance on the validation set. More implementation details can be found in the Appendix.




\begin{table}[h]
    \centering
    \resizebox{\linewidth}{!}{
    \begin{tabular}{|c|R{1.5cm}|R{1.5cm}|R{1.5cm}|}
    \hline
    Model & P & R & F1 \\
    \hline
    \hline
    \texttt{RULE-SVM} &71.09&58.39&64.12 \\
    \texttt{MaxEnt-SVM} &74.99&64.31&69.24 \\
    \texttt{CRF-SVM} &72.27&66.81&69.43\\
    \hline
    \hline
    \texttt{RNN-ATT} &71.96&69.15&70.53\\
    \texttt{SP-ILP} &78.15&\textbf{78.29}&78.22\\
    \texttt{CTRL} &84.88&73.28&78.65 \\
    \hline
    \hline
    \modelname &\textbf{86.80}&74.53&\textbf{80.20} \\
    \hline
    \end{tabular}}
    \caption{Performance of temporal relation extraction on I2B2-2012 datasets. All improvements of \modelname over baseline methods are statistically significant at a 99\% confidence level in paired \textit{t}-tests. Results show that \modelname outperforms all the baselines. }
    \label{tab:result1}
\end{table}
\begin{table}[t]
    \centering
    \resizebox{\linewidth}{!}{
    \begin{tabular}{|l|R{1.4cm}|R{1.4cm}|R{1.4cm}|R{1cm}|}
    \hline
    Feature  & P & R & F1 & Lift\\
    \hline 
    \hline
    Best & 86.80 & 74.53 & 80.20 & -\\
    \hline
    w/o PSL & 85.78 & 73.31 & 79.06 & \textbf{1.44}\% \\
    \hline
    w/o GTI & 85.08 & 73.31 & 78.76 & \textbf{1.83}\% \\
    \hline
    \end{tabular}}
    \caption{Ablation study on I2B2-2012 dataset. GTI denotes the global temporal inference. Results show significant performance lifts from both PSL and GTI modules.}
    \label{tab:abl}
\end{table}
\begin{table}[h]
    \centering
    \resizebox{\linewidth}{!}{
    \begin{tabular}{|l|R{1.3cm}|R{1.3cm}|R{1.3cm}|R{1cm}|}
    \hline
    Strategy  & P & R & F1 & Lift\\
    \hline 
    \hline
     Random & 85.08 & 73.93 & 79.21 & -\\
    \hline
     Confidence & 86.07 & 73.76 & 79.44  & \textbf{0.29\%}\\
    \hline
     Confidence +  & \multirow{2}{*}{86.80} & \multirow{2}{*}{74.53} & \multirow{2}{*}{80.20} & \multirow{2}{*}{\textbf{1.25}\%}\\ 
    Time Anchor & & & &\\
    \hline
    \end{tabular}}
    \caption{Comparison of different ranking methods applied in the global inference on I2B2-2012 dataset.}
    \label{tab:gti}
\end{table}

\begin{table*}[h]
    \centering
    \resizebox{\linewidth}{!}{
    \begin{tabular}{|c|c|c|c|c|}
    \hline
     \multirow{7}{*}{1} & \multirow{2}{*}{Text} & \multicolumn{3}{l|}{{\color{red} Her acute bradycardic event} was felt likely secondary to {\color{blue}her new beta blocker} in conjunction with} \\ 
     & & \multicolumn{3}{l|}{a vagal response . It was determined to stop {\color{green}the beta blocker} , and atropine was placed at the bedside .}\\
     \cline{2-5}
     & (e1, e2) & ({\color{red}Her...event}, {\color{blue}her...blocker}) & ({\color{blue}her...blocker}, {\color{green}the beta blocker}) & ({\color{red}Her...event}, {\color{green}the beta blocker})\\
     \cline{2-5}
     & True Label & After & Overlap & After\\
     \cline{2-5}
     & \texttt{CRTL} & After & Overlap & Overlap \\
     \cline{2-5}
     & \modelname & After & Overlap & After \\
     \cline{2-5}
     &Rule & \multicolumn{3}{c|}{After$(A,B)$ $\land$ Overlap$(B,C)$ $\rightarrow$ After$(A,C)$}\\
    \hline 
    \hline
     \multirow{7}{*}{2} & \multirow{2}{*}{Text} & \multicolumn{3}{l|}{The patient was given an aspirin and Plavix and in addition {\color{red}started} on {\color{blue}a beta Elmore} , {\color{green}Maxine ACE}} \\ 
     & & \multicolumn{3}{l|}{{\color{green}inhibitor} , and these were titrated up as her blood pressure tolerated .}\\
     \cline{2-5}
     & (e1, e2) & ({\color{red}started}, {\color{blue}a beta Elmore}) & ({\color{blue}a beta Elmore}, {\color{green}Maxine ACE}) & ({\color{red}started}, {\color{green}Maxine ACE})\\
     \cline{2-5}
     & True Label & Before & Overlap & Before\\
     \cline{2-5}
     & \texttt{CRTL} & After & Overlap & Before \\
     \cline{2-5}
     & \modelname & After & Overlap & After \\
     \cline{2-5}
     &Rule & \multicolumn{3}{c|}{After$(A,B)$ $\land$ Overlap$(B,C)$ $\rightarrow$ After$(A,C)$}\\
    \hline     
    \hline 
    \multirow{6}{*}{3} & Text & \multicolumn{3}{l|}{She has had attacks treated with {\color{red}antibiotics} in the past notably in {\color{blue}12/96} and {\color{green}08/97} .} \\
     \cline{2-5}
     & (e1, e2) & ({\color{red}antibiotics}, {\color{blue}12/96}) & ({\color{blue}12/96}, {\color{green}08/97}) & ({\color{red}antibiotics}, {\color{green}08/97})\\
     \cline{2-5}
     & True Label & Overlap & Before & Overlap\\
     \cline{2-5}
     & \texttt{CRTL} & Overlap & Before & Overlap \\
     \cline{2-5}
     & \modelname & Overlap & Before & Before \\
     \cline{2-5}
     &Rule & \multicolumn{3}{c|}{Overlap$(A,B)$ $\land$ Before$(B,C)$ $\rightarrow$ Before$(A,C)$}\\
    \hline
    \end{tabular}}
    \caption{Case study and error analysis of the model predictions on I2B2-2012 Dataset.}
    \label{tab:error}
\end{table*}

\begin{table}[t]
    \centering
    \resizebox{\linewidth}{!}{
    \begin{tabular}{|c|r|r|r|r|r|r|}
    \hline
    \multirow{2}{*}{} & \multicolumn{3}{c|}{\texttt{SP-ILP}} & \multicolumn{3}{c|}{\modelname} \\
    \cline{2-7}
    & P & R & F1 & P & R & F1 \\
    \hline
    \hline
    Before &71.1&58.9&64.4&52.6&74.8&61.7 \\
    After &75.0&55.6&63.5&69.0&72.5&70.7 \\
    Includes &24.6&4.2&6.9&60.9&29.8&40.0 \\
    Is\_Include &57.9&5.7&10.2&34.7&27.7&30.8 \\
    Simultaneous &-&-&-&-&-&-\\
    Vague &58.3&81.2&67.8&72.8&64.8&68.6 \\
    \hline
    Micro-average  &\multicolumn{3}{r|}{63.2}&\multicolumn{3}{r|}{\textbf{65.2}}  \\
    \hline
    \hline
    \texttt{CAEVO} &\multicolumn{6}{r|}{49.4} \\
    \texttt{LSTM-DP} &\multicolumn{6}{r|}{52.9} \\
    \texttt{GCL} &\multicolumn{6}{r|}{57.0} \\
    \texttt{CTRL} &\multicolumn{6}{r|}{63.6} \\
    \hline
    \end{tabular}}
    \caption{Performance of temporal relation extraction on TB-dense datasets. All improvements of \modelname over baseline methods are statistically significant at a 99\% confidence level in paired \textit{t}-tests. We also compare the breakdown performance for each relation class between \modelname and \texttt{SP-ILP}.}
    \label{tab:result2}
\end{table}

\subsection{Experimental Results}
Table~\ref{tab:result1} and Table~\ref{tab:result2} contains our main results. As we observe, our \modelname enhanced by PSL regularization and global inference achieve the best relation extraction performances per F1 score. Compared with the baseline models, the F1 score improvements are 2.0\% and 2.5\% on I2B2-2012 and TB-Dense data respectively, which are all statistically significant.
\\

\noindent \textbf{I2B2-2012.}
As shown in Table~\ref{tab:result1}, our model \modelname outperforms the best baseline method \texttt{CTRL} by 2\% and outperforms the structured prediction method \texttt{SP-ILP} by 2.5\% per F1 score. \texttt{SP-ILP} gets the highest Recall score, but sacrifice the predicting precision instead. We also observe that by simply fine-tuning the BERT to generate the sentence embeddings and then feeding them into one layer of FFN for classification, \texttt{CTRL} can achieve an impressive F1 score of 78.65\%. This proves the advantage of contextualized embeddings over static embeddings used by other baseline models. Besides, \modelname outperforms the feature-based systems, \texttt{CRF-SVM} and \texttt{MaxEnt-SVM}, by over 10\% per F1 score. 

We develop an ablation study to test different features, as shown in Table~\ref{tab:abl}. We see that PSL regularization and global temporal inference modules lift the performance by 1.44\% and 1.83\% separately. Both Precision and Recall performances are improved. We can clearly conclude that learning the relations with the proposed algorithms improves our model significantly (also at a 99\% level in paired \textit{t}-tests). 

We also show the comparisons among different ranking strategies for the global inference module in Table~\ref{tab:gti}. Random denotes that we randomly add a new prediction to the time graph and resolve the conflict. Confidence denotes we rank the predictions per the prediction probabilities and then add them to the graph in decreasing order. Time Anchor represents that we first construct the time graph based on the predictions for temporal relations of type \texttt{T-T}. In the results, we see a 0.29\% improvement per F1 score when switching from the Random to the Confidence strategy. After adding the Time Anchor method, we observe a 1.25\% performance lift, compared to Random strategy. 
This proves the effectiveness of the time-anchored global temporal inference module. 
\\

\noindent \textbf{TB-Dense.}
We show the experimental results on TB-Dense dataset in Table~\ref{tab:result2}. Our model outperforms the best baseline model \texttt{CTRL} by 2.5\% and outperforms the structured prediction method \texttt{SP-ILP} by 3.2\% per Micro-average F1 score. We observe that in the performance breakdown for each relation class, \modelname obtains similar scores on \underline{Before}, \underline{After}, and \underline{Vague} as \texttt{SP-ILP} and gets much better performances on \underline{Is\_Include} and \underline{Includes}. These two types only occupy 5.7\% and 4.5\% of all the instances. \modelname and \texttt{SP-ILP} both fail to label any instance as \underline{Simultaneous} because of its even fewer instances (1.5\%) for training.

Besides, we observe \modelname achieves higher Recall values in all the categories of temporal relations, which prove that incorporating the dependency rules into model training can dramatically lift the coverage of predictions.

\subsection{Case Study and Error Analysis.}
Table~\ref{tab:error} shows the results of a case study with the outputs of \texttt{CTRL} and \modelname. In the first case, the temporal relation between \textit{Her acute bradycardic event} and \textit{the beta blocker} is hard to predict due to the noise brought by the long context. \texttt{CTRL} predicts it as \underline{Overlap}, while \modelname corrects it to \underline{After} according to the potential PSL rule that can be matched with the first two correct predictions. In some cases, however, \modelname will make new mistakes. For example in case 2, if our model initially predicts the relation between \textit{started} and \textit{a beta Elmore} wrong, a potential PSL rule sometimes will lead to an extra mistake when predicting the relation between \textit{started} and \textit{Maxine ACE}. In the case 3, \textit{antibiotics} treated the \textit{attacks} twice in both $12/96$ and $08/97$, where the PSL rule is no longer valid since the \textit{antibiotics} in fact denote two occurrences of this event. In such special cases with invalid rules, \modelname may make a mistake.  
\section{Related Work}
\subsection{Clinical Temporal Relation Extraction}
\noindent \textbf{Corpora. } 
Different from the datasets in the news domain~\cite{pustejovsky2003timebank,AQUAINT}, the corpora in the clinical domain require rich domain knowledge for annotating the temporal relations.
I2b2-2012~\cite{sun2013evaluating} and Clinical TempEval~\cite{bethard-etal-2015-semeval,bethard-etal-2016-semeval,bethard-etal-2017-semeval} are some great efforts of building clinical datasets with extensive annotations including labels of clinical events and temporal relations, the second of which was not tested in our paper due to lack of access to the data.

\noindent \textbf{Models.} 
Some early efforts to solve the clinical relation extraction problem leverage conventional machine learning methods~\cite{llorens2010tipsem,sun2013evaluating,xu2013end,tang2013hybrid,lee-etal-2016-uthealth,chikka-2016-cde} such as SVMs, MaxEnt and CRFs, and neural network based methods~\cite{lin2017representations,lin2018self,dligach2017neural,tourille2017neural,lin-etal-2019-bert,guan2020robustly,lin-etal-2020-bert,galvan2020empirical}. They either require expensive feature engineering or fail to consider the dependencies among temporal relations within a document. \cite{leeuwenberg-moens-2017-structured,han-etal-2019-deep,han-etal-2019-joint,ning-etal-2017-structured} formulate the problem as a structured prediction problem to model the dependencies but can not globally predict temporal relations. Instead, our method can infer the temporal relations at document level.

\subsection{Probabilistic Soft Logic}
In recent years, PSL rules have been applied to various machine learning topics such as Fairness~\cite{farnadi2019declarative}, Model Interpretability~\cite{hu-etal-2016-harnessing}, Probabilistic Reasoning~\cite{augustine2019tractable,dellert2020exploring}, Knowledge Graph Construction~\cite{pujara2013knowledge,chen2019embedding} and Sentiment Analysis~\cite{deng-wiebe-2015-joint,gridach2020framework}. We are the first to model the temporal dependencies with PSL. 
\section{Conclusion}

In this paper, we propose \modelname that leverages the PSL rules to model the temporal dependencies as a regularization term to jointly learn a relation classification model. Extensive experiments show the efficacy of the PSL regularization and global temporal inference with time graphs. 

\section{Acknowledgement}
We would like to thank the anonymous reviewers for their helpful comments. The work was supported by NSF DBI-1565137, DGE-1829071, NIH R35-HL135772, NSF III-1705169, NSF CAREER Award 1741634, NSF \#1937599, DARPA HR00112090027, Okawa Foundation Grant, and Amazon Research Award.

{ \small
\bibliography{aaai21}
}
\appendix
\clearpage
\newpage

\section{Appendices}
\label{sec:appendix}
\subsection{Dataset}
\begin{figure}[ht]
    \centering
     \begin{subfigure}[b]{\linewidth}
     \centering
    \includegraphics[width=\linewidth]{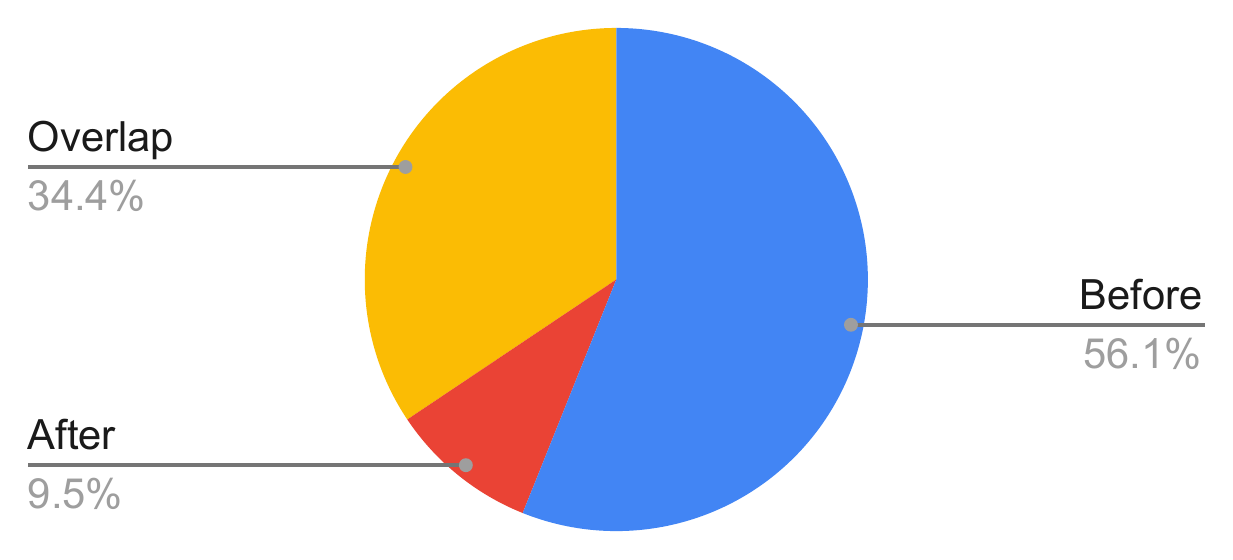}
    \caption{I2B2-2012}
    \label{fig:i2b2data}

    \end{subfigure}
    \begin{subfigure}[b]{\linewidth}
    \centering
    \includegraphics[width=\linewidth]{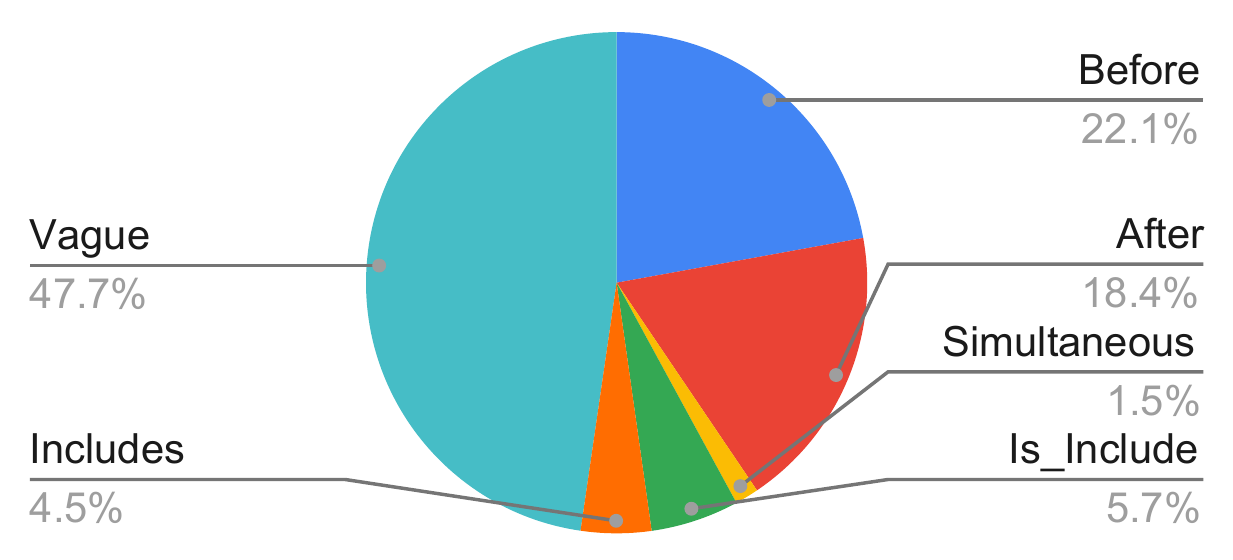}
    \caption{TB-Dense
    }
    \label{fig:tbddata}

    \end{subfigure}

\caption{Relation Type Distribution of I2B2-2012 and TB-Dense}
\label{fig:type}
\end{figure}

The I2B2-2012 dataset\footnote{\url{https://www.i2b2.org/NLP/DataSets/}} is officially split into training and test sets, containing 190 and 120 documents, separately. We randomly sampled 5\% of training data as a validation set. 
For 
TB-Dense\footnote{\url{https://www.usna.edu/Users/cs/nchamber/caevo/}}, the training/validation/test sets are given. The statistics are shown in Table~\ref{tab:data} and we plot the detailed relation type distributions of two datasets in Figure~\ref{fig:type}. We observe that 
TB-Dense is a relatively unbalanced dataset, where \underline{Vague} dominates the dataset.  We compute the density by (\# existing relations)/(\# possible pairs of entities).
Following \cite{han-etal-2019-joint,meng-rumshisky-2018-context}, we only predict the \texttt{E-E} relations and exclude all other relations from evaluation for TB-Dense dataset evaluation. Therefore, we cannot apply the global inference module to this dataset.

\subsection{Data Preprocessing}
To facilitate the PSL rule grounding and distance calculation, 
we arrange the training data as a collection of instances, each of which contains three pairs of relations. 
We first traverse the training dataset to match the PSL rules and make sure that each ground rule is treated as one instance to further calculate the PSL loss term. Besides, relations that are not involved in any rules are packed together and we do not have to compute a PSL loss for them.
We also augment the training data by flipping every pair according to the symmetry rules, i.e. if \underline{Before}($A,B$) is $true$, \underline{After}($B,A$)  should also be $true$. We incorporate the new relations into the training dataset to alleviate the unbalanced data issue. 

\subsection{Model Training Details}
We leverage the pretrained BERT-BASE model~\cite{devlin2018bert} to generate the sentence embeddings, which contains 110M parameters to fine-tune. 
In the experiments, we save the checkpoint with the highest validation performance for final testing. In Table~\ref{tab:dev}, we list the best validation performance of different datasets and the corresponding test performance that we also reported in Table~\ref{tab:result1} and Table~\ref{tab:result2} for completeness.
Data and codes are attached. 

\begin{table}[h]
    \centering
    \resizebox{.95\linewidth}{!}{
    \begin{tabular}{|c|l|r|r|r|}
    \hline
        \multicolumn{2}{|c|}{Dataset} & Precision & Recall & F1 \\
        \hline
        \hline
        \multirow{2}{*}{I2B2-2012} & Test & 86.80 & 74.53 & 80.20 \\
        \cline{2-5}
        & Val & 86.03 & 86.03 & 86.03 \\ 
        \hline
        \multirow{2}{*}{TB-Dense} & Test & 65.20 & 65.20 & 65.20 \\
        \cline{2-5}
        & Val & 58.80 & 58.80 & 58.80 \\  
        \hline
    \end{tabular}}
    \caption{Corresponding validation performance for each reported test result of \modelname in Table~\ref{tab:result1} and Table~\ref{tab:result2}.}
    \label{tab:dev}
\end{table}


\end{document}